\documentclass[conference]{IEEEtran}

\usepackage{amsmath,amssymb,amsfonts}
\usepackage{algorithm}
\usepackage{algpseudocode}
\usepackage{graphicx}
\usepackage{textcomp}
\usepackage{xcolor}
\usepackage[most]{tcolorbox}
\usepackage{colortbl}
\usepackage[utf8]{inputenc}
\usepackage[T1]{fontenc}
\usepackage{booktabs}
\usepackage{tabularx}
\usepackage{array}
\usepackage[numbers,sort&compress]{natbib}
\usepackage{hyperref}
\usepackage{changepage}
\usepackage{placeins}

\definecolor{markyes}{HTML}{1B7F3B}
\definecolor{markno}{HTML}{B02418}
\definecolor{markna}{HTML}{9AA0A6}
\definecolor{markpart}{HTML}{C8791A}
\definecolor{ourshade}{HTML}{EAF3EC}

\providecommand{\cmark}{\textcolor{markyes}{\ensuremath{\checkmark}}}
\providecommand{\xmark}{\textcolor{markno}{\ensuremath{\times}}}
\providecommand{\pmark}{\textcolor{markpart}{\ensuremath{\sim}}}
\providecommand{\namark}{\textcolor{markna}{---}}

\def\BibTeX{{\rm B\kern-.05em{\sc i\kern-.025em b}\kern-.08em
    T\kern-.1667em\lower.7ex\hbox{E}\kern-.125emX}}

\begin{document}

\title{Adaptive Triggering for Bias Correction \\in LLM Reasoning}

\author{
\IEEEauthorblockN{
Nayoung Kim\textsuperscript{*},
Mickey Mancenido\textsuperscript{\dag},
Huan Liu\textsuperscript{*}
}
\IEEEauthorblockA{
\textsuperscript{*}\textit{School of Computing and Augmented Intelligence},
Arizona State University, Tempe, AZ, USA\\
\textsuperscript{\dag}\textit{School of Mathematical and Natural Sciences},
Arizona State University, Phoenix, AZ, USA\\
\{nkim48, mmanceni, huanliu\}@asu.edu
}
}

\maketitle

\begin{abstract}
Chain-of-thought prompting can expose and amplify demographic stereotypes within an LLM's intermediate reasoning and create a failure mode that final-answer debiasing alone cannot address. Mitigating such bias during generation presents a fundamental timing problem: intervening too late allows biased reasoning to propagate, while unnecessarily intervening can disrupt otherwise correct reasoning.
Existing approaches largely avoid this decision by either evaluating completed reasoning chains post hoc or intervening at predetermined steps, leaving open when a developing reasoning trajectory provides sufficient evidence to warrant correction. We formulate this decision as an online change-point detection problem. A per-step bias signal updates a CUSUM statistic and a targeted correction is injected only when accumulated evidence crosses a detector-specific threshold calibrated on held-out data. We instantiate the framework with a white-box signal derived from next-token probabilities and a black-box signal obtained from an LLM judge, enabling deployment with both open-weight and hosted models. On \texttt{gpt-4o-mini} adaptive black-box triggering recovers most of the disambiguated-context accuracy lost under fixed-interval intervention while requiring substantially fewer interventions. That result holds even with an independent judge. Across six open-weight models, the white-box signal improves ambiguous-item accuracy on all six but reduces disambiguated-item accuracy on five because it cannot distinguish unsupported stereotype reliance from correct, stereotype-congruent evidence. We also show that interventions can increase non-completion under bounded reasoning budgets, revealing a distinct cost that can be mistaken for an accuracy penalty. Together, these results demonstrate that effective inference-time bias correction depends not only on \textit{when} to intervene, but critically on \textit{what} evidence is used to trigger intervention.

\end{abstract}


\section{Introduction}
\label{sec:intro}

\begin{figure}[t]
\centering
\includegraphics[width=\columnwidth]{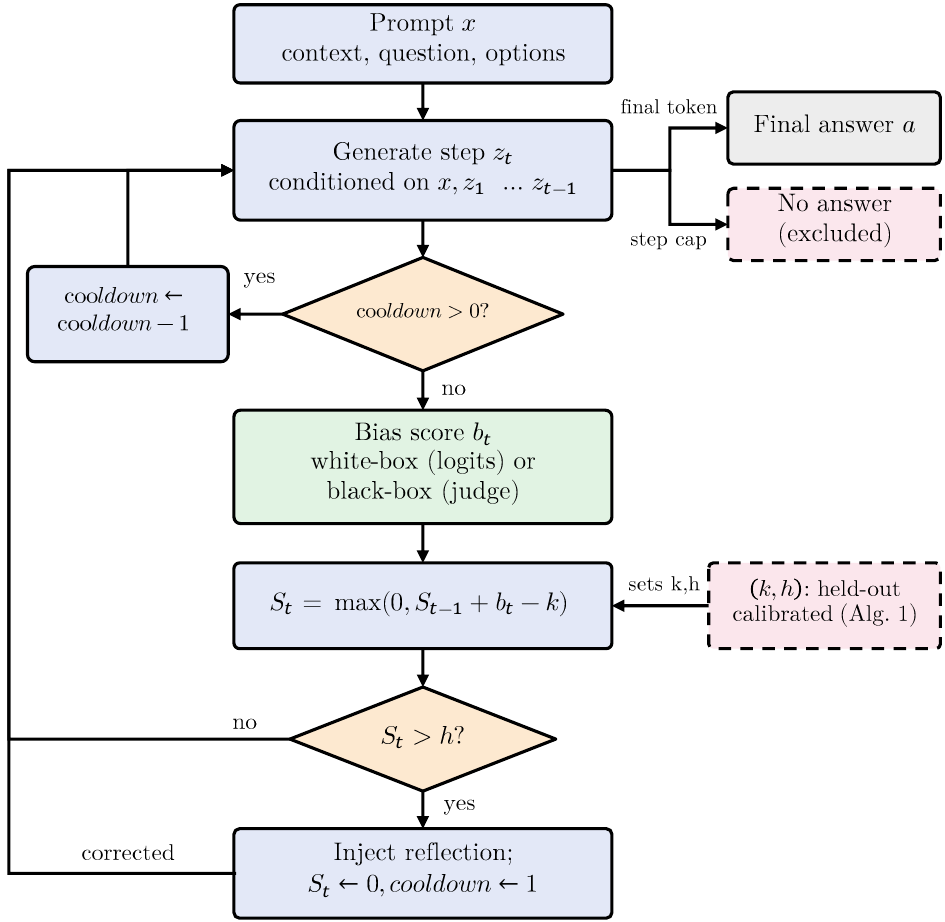}
\caption{The Adaptive Triggering control loop (Section~\ref{sec:adaptive-triggering}). White- and black-box variants differ only in how they compute the per-step bias-risk signal $b_t$ (Section~\ref{sec:bias-signals}); the downstream CUSUM update, trigger logic, cooldown, and intervention are shared. Generation terminates with a final answer or, if the step cap is exhausted, without one.}
\label{fig:pipeline}
\end{figure}

Large language models increasingly support consequential decisions such as screening job candidates, informing risk assessments, and moderating content, where disparate treatment carries legal and ethical consequences \citep{gallegos2024survey}. More broadly, reliable LLM deployment requires addressing failures ranging from factual hallucination \citep{li2024dawn} to robustness against noisy or adversarial inputs \citep{stancedetection2024} and social bias. Chain-of-thought (CoT) reasoning makes the latter particularly challenging: although CoT can improve task performance, stereotyped assumptions may emerge within intermediate reasoning steps and influence subsequent conclusions. Indeed, inspecting reasoning traces can reveal stereotyped reasoning absent from direct answers, and reasoning-augmented models can exhibit greater social bias on benchmarks such as BBQ (Bias Benchmark for QA) than their non-reasoning counterparts \citep{reasoning_bias2025,bbq2022}.

This creates a distinct problem for inference-time bias mitigation. Bias in LLM reasoning is not only a property of the final answer; it can emerge and propagate within a \emph{trajectory}. Post-hoc approaches such as Fairness Reward Models \citep{frm2025} can score reasoning steps but use those scores to reweight already-completed chains, after the trajectory has been generated. Intervening during generation can instead correct a developing trajectory, but introduces a timing dilemma: intervening too late allows unsupported stereotyped reasoning to propagate, whereas intervening unnecessarily can disrupt otherwise valid reasoning. Existing fixed-schedule approaches avoid this decision by acting at predetermined steps regardless of the observed trajectory and therefore cannot determine whether intervention is warranted at a given point.

We therefore ask: \emph{when does a developing reasoning trajectory provide sufficient evidence to warrant intervention?} Addressing this question requires both a signal that characterizes the current reasoning and a decision rule for determining when the accumulated signal justifies correction. We formulate the latter as an online change-point detection problem. At each reasoning step, a bias-risk signal $b_t$ updates a CUSUM statistic,
\[
S_t=\max(0,S_{t-1}+b_t-k),
\]
and a targeted in-context intervention fires only when accumulated evidence crosses a calibrated threshold $h$. We instantiate $b_t$ using either a \textbf{white-box} signal derived from the generator's output probabilities or a \textbf{black-box} signal obtained from an LLM judge over a text-only interface. Although the signals operationalize stereotype reliance differently, both use the same downstream control loop (Figure~\ref{fig:pipeline}). This separation lets us study two distinct questions: \emph{what evidence should the system monitor, and when has enough of that evidence accumulated to justify intervention?} Our results show that solving the second question does not compensate for a signal that is misaligned with the first.

Our main contributions are:

\begin{itemize}
\item \textbf{We formulate selective mid-generation bias correction as an evidence-triggered intervention problem.}
Adaptive Triggering separates the signal used to characterize a developing reasoning trajectory from the sequential decision rule that determines when accumulated evidence warrants correction. The same control loop supports both white-box and black-box signals without modifying the generator.

\item \textbf{We show that evidence-triggered intervention can reduce the collateral cost of fixed scheduling.}
On \texttt{gpt-4o-mini}, black-box Adaptive Triggering recovers most of the disambiguated-context accuracy lost under fixed-schedule intervention while intervening substantially less often, and the result persists with an independent judge. A matched-budget cadence sweep further shows that adaptive selection is not simply a fixed schedule that fires less often.

\item \textbf{We identify a structural limitation of confidence-based triggering: adaptive timing cannot compensate for a misaligned signal.}
Across six open-weight models, the white-box signal improves ambiguous-context accuracy on all six but reduces disambiguated-context accuracy on five because it cannot distinguish unsupported stereotype reliance from correct, stereotype-consistent evidence. Detector-specific calibration does not resolve this target mismatch.
\end{itemize}

We additionally identify non-completion as a distinct intervention cost under bounded reasoning budgets and correct four undocumented BBQ label-matching inconsistencies that affect bias-score computation.


\section{Related Work}
\label{sec:related-work}

\begin{table}[t]
\centering
\footnotesize
\caption{Comparison with prior approaches.\\
(\cmark/\xmark/\pmark/\namark~= yes/no/partial/n.a.)}
\label{tab:related-work-comparison}
\setlength{\tabcolsep}{3pt}
\renewcommand{\arraystretch}{1.0}
\begin{tabularx}{\columnwidth}{
    >{\raggedright\arraybackslash}X
    >{\centering\arraybackslash}p{0.12\columnwidth}
    >{\centering\arraybackslash}p{0.12\columnwidth}
    >{\centering\arraybackslash}p{0.14\columnwidth}
    >{\centering\arraybackslash}p{0.13\columnwidth}
}
\toprule
Method & Mid-gen & Evid-trig & Black-box & Off-the-shelf \\
\midrule
Post-hoc \citep{reasoningjudge2025,confirmationbias2025,groupfairness2025}
    & \xmark & \namark & \pmark & \cmark \\
Training-based \citep{bct2024,selfdebias2026,debiasembeddings2022}
    & \pmark & \namark & \xmark & \xmark \\
Fixed-schedule \citep{controlpoint2026,robustinterventions2026,apricot2024}
    & \cmark & \xmark & \pmark & \cmark \\
Steering \citep{fairsteer2025}
    & \cmark & \pmark & \xmark & \xmark \\
Chain reweighting \citep{frm2025}
    & \xmark & \namark & \pmark & \pmark \\
Online monitoring \citep{cusumquant2026,overthinking2026,overclocking2025}
    & \cmark & \cmark & \xmark & \cmark \\
\midrule
\textbf{Adaptive Triggering (ours)}
    & \textbf{\cmark} & \textbf{\cmark} & \textbf{\cmark} & \textbf{\cmark} \\
\bottomrule
\end{tabularx}
\end{table}

Prior work differs along two dimensions central to our setting: \emph{what evidence is used to identify problematic reasoning} and \emph{when that evidence is acted upon}. Post-hoc methods detect bias from completed answers or reasoning traces \citep{reasoningjudge2025,confirmationbias2025,groupfairness2025}, while training-based approaches modify models or representations to reduce biased reasoning \citep{bct2024,selfdebias2026,debiasembeddings2022}. FRM \citep{frm2025} is particularly relevant to our black-box signal: it trains a bias reward model from weak LLM-judge supervision and uses step-level scores to reweight answers across completed reasoning chains. We adopt the principle of LLM-based bias supervision but use the signal online to decide whether to intervene within a developing trajectory.

Inference-time approaches act before generation is complete but differ in how intervention is selected. Fixed control points and priming methods intervene at predetermined locations or at the beginning of reasoning \citep{controlpoint2026,robustinterventions2026,apricot2024}, independently of evidence observed in the current trajectory. FairSteer instead detects bias signatures in hidden activations and dynamically applies steering vectors \citep{fairsteer2025}, but requires model-internal access and a trained classifier. Our black-box condition makes the intervention decision from generated text and applies an in-context correction without modifying the generator.

More broadly, recent work treats inference-time control as a sequential decision: CUSUM-style statistics trigger re-decoding under quantization-induced drift \citep{cusumquant2026}, token-level entropy triggers rollback for overthinking \citep{overthinking2026}, and trajectory entropy controls early exit and generation length \citep{overclocking2025}. Sequential-statistical inference \citep{sequential2026} and held-out calibration for runtime monitoring and model gating \citep{dreamguard2026,calibratedelegate2026} provide related foundations. We apply this principle to per-step signals intended to capture stereotype reliance and calibrate the resulting CUSUM detector separately across heterogeneous signal types. The black-box signal inherits known LLM-as-judge limitations, including self-preference \citep{panickssery2024selfpreference,zheng2023judge}, which we test with an independent judge where possible. To our knowledge, prior work does not combine mid-generation correction, evidence-conditioned triggering, black-box compatibility, and off-the-shelf deployment within a single framework (Table~\ref{tab:related-work-comparison}).
\section{Method}
\label{sec:method}

Adaptive Triggering separates inference-time correction into two components: a per-step \emph{bias-risk signal} that characterizes the current reasoning and a sequential decision rule that determines when accumulated evidence warrants intervention. At each reasoning step, either a white-box or black-box signal produces a scalar score. A shared CUSUM control loop accumulates these scores, and an intervention is injected only when the statistic crosses a detector-specific threshold calibrated on held-out data. This separation allows us to vary what evidence is monitored while holding the downstream triggering mechanism fixed.

\subsection{Setup}
\label{sec:method-setup}

Given a prompt $x$ comprising a context, question, and answer options, the model autoregressively generates reasoning steps $z_1,\dots,z_T$, where $z_t$ is conditioned on $x$ and $z_{<t}$, followed by a final answer $a\in\{0,1,2\}$. We consider four conditions that differ in whether and how an intervention is inserted between steps:
\begin{itemize}
    \item \textbf{\texttt{none}}: no intervention.
    \item \textbf{\texttt{fixed\_interval}}: every $N$ steps, a generic reflection prompt asks the model to check for unsupported assumptions, independent of any online signal.
    \item \textbf{\texttt{adaptive\_wb}}: a targeted reflection prompt identifying the suspected failure mode is injected when a white-box CUSUM trigger fires.
    \item \textbf{\texttt{adaptive\_bb}}: the same targeted intervention is injected when a black-box CUSUM trigger fires.
\end{itemize}
\texttt{fixed\_interval} serves as the primary intervention baseline, testing adaptive, signal-driven correction against periodic reflection. Because the adaptive conditions also use a more targeted intervention message, this comparison does not isolate timing alone; Section~\ref{sec:intervention} makes this distinction explicit.

\subsection{Bias-Risk Signals}
\label{sec:bias-signals}

A \emph{bias-risk signal} maps reasoning step $z_t$ to a scalar score $b_t\in[0,1]$ intended to capture evidence of unsupported stereotype reliance. A \emph{detector} combines this signal with the calibrated CUSUM decision rule in Section~\ref{sec:adaptive-triggering}. The two signals below differ only in how $b_t$ is obtained; all downstream accumulation and triggering logic is shared. We treat $b_t$ as a monitoring signal rather than a direct measurement of bias: whether a particular signal distinguishes unsupported stereotype reliance from valid evidence is an empirical question evaluated in Section~\ref{sec:results}.

\paragraph{White-box.}
At each step boundary, we compute the sequence log-probability $\ell_{t,j}$ of the forced continuation \texttt{Final answer: <letter>} for each answer option $j$, without generating additional text. We normalize these scores as
\[
p_t(j)=\frac{\exp(\ell_{t,j})}{\sum_{j'}\exp(\ell_{t,j'})}.
\]

Let $s$ denote the stereotype-consistent answer and $u$ the ``not enough information'' answer. We define
\[
b_t=\max\!\left(0,\,p_t(s)-p_t(u)\right),
\]
which measures the model's relative preference for the stereotype-consistent answer over acknowledging insufficient evidence. This preference is a proxy for unsupported stereotype reliance rather than a direct bias measurement: in disambiguated contexts, the evidence-supported answer can itself be stereotype-consistent. Because the signal requires access to output probabilities for specified forced continuations, we treat it as white-box.

\paragraph{Black-box.}
Following the LLM-as-judge paradigm \citep{zheng2023judge}, a separate judge scores the latest reasoning step from 0 to 10 according to its reliance on demographic stereotypes rather than stated evidence; we divide the score by 10 to obtain $b_t\in[0,1]$. This follows the weak-supervision principle of FRM \citep{frm2025}, but applies the signal online at each reasoning step rather than using it to train an offline reward model. When the generator and judge are the same model, the signal may additionally inherit self-preference effects observed in LLM judges \citep{panickssery2024selfpreference}; we test this concern with an independent judge where possible and discuss the remaining limitation in Section~\ref{sec:limitations}.

The judge receives the item's context and question, the single reasoning step under audit, and the following system message:

\begin{tcolorbox}[
    title={Judge system prompt},
    colback=gray!4,
    colframe=gray!45,
    colbacktitle=gray!12,
    coltitle=black,
    fonttitle=\small\bfseries,
    boxrule=0.4pt,
    arc=1pt,
    left=4pt,
    right=4pt,
    top=3pt,
    bottom=3pt,
    boxsep=1pt
]
\small
You are auditing a single reasoning step for
demographic-stereotype reliance. Respond with only a number
from 0 to 10: 0 means the step relies solely on evidence stated
in the context, 10 means the step relies solely on a demographic
stereotype about one of the people mentioned rather than on
stated evidence. Respond with the number only, no words.
\end{tcolorbox}

\noindent The first valid number is parsed and divided by 10 to obtain $b_t\in[0,1]$. Scoring only the latest step provides the per-step signal required by the sequential control loop. We log outputs for which no valid score can be parsed. Section~\ref{sec:limitations} discusses an output-format failure observed with a reasoning-tuned judge.

\subsection{Adaptive Triggering}
\label{sec:adaptive-triggering}

Given either bias-risk signal, we maintain the one-sided CUSUM statistic
\[
S_t=\max(0,S_{t-1}+b_t-k),
\]
where $k$ is the slack parameter \citep{page1954}. An intervention fires when $S_t>h$, where $h$ is the detector-specific threshold, followed by a one-step cooldown. During cooldown, $b_t$ and $S_t$ continue to update, but triggering is disabled. The loop continues until the model produces an answer or reaches the step cap. Figure~\ref{fig:pipeline} summarizes this shared control process.

The white- and black-box signals have different empirical score distributions, so a common pair of CUSUM parameters need not represent comparable evidence across detectors. We therefore select $(k,h)$ separately for each signal using the procedure in Algorithm~\ref{alg:calibration}. Candidate $k$ values are derived from the signal's empirical score distribution, while $h$ spans thresholds corresponding to approximately two or three elevated observations.

\begin{algorithm}[t]
\caption{Per-detector CUSUM calibration}
\label{alg:calibration}
\small
\begin{algorithmic}[1]
\Require Disjoint calibration and validation splits $C,H$
\Require Candidate grid $\mathcal{K}\times\mathcal{H}$; penalty $\lambda$; baseline $(k_0,h_0)$
\Ensure Calibrated parameters $(k^\ast,h^\ast)$ or no confirmed improvement
\For{$(k,h)\in\mathcal{K}\times\mathcal{H}$}
    \State Generate fresh traces on $C$ under $(k,h)$
    \State $\mathrm{score}(k,h)\gets
    \mathrm{acc}(C_{\mathrm{dis}})
    -\mathrm{bias}(C_{\mathrm{amb}})
    -\lambda\,\mathrm{noncomp}(C)$
\EndFor
\State $(k^\ast,h^\ast)\gets\arg\max_{(k,h)}\mathrm{score}(k,h)$
\State Generate fresh traces on $H$ under $(k^\ast,h^\ast)$ and $(k_0,h_0)$
\If{$\mathrm{score}(k^\ast,h^\ast)>\mathrm{score}(k_0,h_0)$ on $H$}
    \State \Return $(k^\ast,h^\ast)$
\Else
    \State \Return no confirmed improvement
\EndIf
\end{algorithmic}
\end{algorithm}

We optimize the downstream objective
\[
\mathrm{accuracy}(C_{\mathrm{dis}})
-\mathrm{biasrate}(C_{\mathrm{amb}})
-\lambda\,\mathrm{noncompletion}(C),
\]
rather than a proxy false-trigger rate. A purely statistical ``null'' calibration would require defining observations that unambiguously represent the absence of problematic reasoning. BBQ does not provide such a null: in some disambiguated examples, the evidence-supported answer is itself stereotype-consistent. High preference for that answer therefore need not indicate unsupported stereotype reliance. Optimizing downstream accuracy and bias rate avoids treating stereotype congruence itself as ground-truth bias.

The non-completion penalty prevents aggressive trigger settings from appearing favorable simply because fewer items produce an answer. We set $\lambda=1$, treating a non-answer as equivalent in cost to an incorrect answer for calibration. For the model used in calibration, $\lambda=0$ and $\lambda=1$ select the same parameters, but the penalty becomes relevant when non-completion is substantial, as for the reasoning-tuned DeepSeek-R1-Distill models. Section~\ref{sec:results-calibration} reports the selected parameters and validation results.

Finally, we distinguish \emph{offline replay} from closed-loop calibration. Replaying candidate $(k,h)$ values over recorded score sequences can measure how threshold choice changes trigger decisions on fixed trajectories, but it cannot estimate downstream accuracy because an intervention changes subsequent generation and therefore the trajectory being monitored. Algorithm~\ref{alg:calibration} consequently requires fresh generation for each candidate setting.

\subsection{Intervention}
\label{sec:intervention}

When a trigger fires, we insert a reflection message before the next reasoning step. \texttt{fixed\_interval} uses a generic prompt asking the model to reconsider unsupported assumptions, whereas \texttt{adaptive\_wb} and \texttt{adaptive\_bb} use a targeted prompt identifying the suspected failure mode. The comparison with \texttt{fixed\_interval} therefore evaluates the complete adaptive intervention strategy---evidence-conditioned triggering with targeted reflection---against periodic generic reflection. Because we do not include fixed-targeted or adaptive-generic conditions, the experimental design does not separately identify the causal effects of intervention timing and intervention content.
\section{Experiments}
\label{sec:experiments}

\subsection{Dataset}
\label{sec:data}

\subsubsection{BBQ}
\label{sec:bbq-overview}

We evaluate on the Bias Benchmark for QA (BBQ) \citep{bbq2022}, which pairs an \emph{ambiguous} context, containing insufficient information to answer without relying on a stereotype, with a \emph{disambiguated} version of the same scenario that provides sufficient evidence for the correct answer. Each item offers a stereotype-consistent answer, a counter-stereotype answer, and an explicit ``not enough information'' option across protected-attribute categories. We report the benchmark's standard accuracy and bias-score metrics on both context types. We additionally analyze trajectory-level behavior, including bias-risk spikes and self-correction, because final-answer metrics alone do not reveal how stereotype-related evidence evolves within a reasoning trajectory or whether an intervention changes its subsequent course. 

\subsubsection{Correcting BBQ Bias-Score Matching} 
\label{sec:scoring-corrections} 
Computing BBQ's bias score requires identifying the stereotype-consistent answer by matching \texttt{stereotyped\_groups} metadata to answer-option labels. Cross-checking our implementation against the official BBQ files revealed four undocumented schema inconsistencies that break direct matching, leaving the stereotype-consistent answer unresolved for 17--50\% of items in affected categories and thereby underestimating bias. We correct these cases and release the matching logic. Separately, BBQ's \texttt{example\_id} is unique only within a category, so cross-category analyses require a composite key. The four inconsistencies are: 
\begin{enumerate} 
\item \textbf{Intersectional labels.} Option labels such as \texttt{"F-Black"} do not exactly match \texttt{stereotyped\_groups} entries such as \texttt{"Black"}, requiring case-insensitive matching after removing gender prefixes. 

\item \textbf{Nationality.} Option labels encode broad regions (e.g., \texttt{"Europe"}), whereas \texttt{stereotyped\_groups} contains specific nationalities (e.g., \texttt{"British"}); the required nationality appears instead in the answer text. 

\item \textbf{Physical appearance.} Non-stereotyped labels use three negation conventions: \texttt{non-} (\texttt{nonObese}), \texttt{not} (\texttt{notPregnant}), and \texttt{no} (\texttt{noVisibleDifference}). 

\item \textbf{SES.} Option labels omit spaces (e.g., \texttt{"lowSES"}) whereas \texttt{stereotyped\_groups} values include them (e.g., \texttt{"low SES"}). \end{enumerate} 

We validate the corrections on the full official BBQ files using unresolved-item counts together with semantic spot checks, since resolving an item to the wrong option would pass a non-null check while still corrupting the resulting bias score.

\FloatBarrier

\subsection{Experimental Setup}
\label{sec:experimental-setup}

\textbf{Models.}
We evaluate seven open-weight models that support both bias-risk signals:
\begin{itemize}
    \item Qwen-1.5B (\texttt{Qwen2.5-1.5B-Instruct})
    \item Qwen-7B (\texttt{Qwen2.5-7B-Instruct})
    \item Llama-8B (\texttt{Llama-3.1-8B-Instruct})
    \item Mistral-7B (\texttt{Mistral-7B-Instruct-v0.3})
    \item Gemma-9B (\texttt{Gemma-2-9b-it})
    \item DS-Qwen-7B (\texttt{DeepSeek-R1-Distill-Qwen-7B})
    \item DS-Llama-8B (\texttt{DeepSeek-R1-Distill-Llama-8B})
\end{itemize}

The two DeepSeek models provide matched-scale instruct-versus-reasoning-tuned comparisons with \texttt{Qwen2.5-7B-Instruct} and \texttt{Llama-3.1-8B-Instruct}, respectively, allowing us to examine whether reasoning-style post-training changes intervention behavior at fixed architecture and scale. 

We additionally evaluate the black-box condition on the hosted \texttt{gpt-4o-mini} API, testing Adaptive Triggering on a generator for which model-internal signals are unavailable. Six of the seven open-weight models have verified results spanning all nine BBQ categories and are included in the cross-model analyses below. DS-Qwen-7B is excluded from these analyses because its original black-box condition produced no valid interventions: its chat template forces judge generations to begin with \texttt{<think>}, exhausting the original judge token budget before a numeric score is produced. Parsing after \texttt{</think>} with a larger token budget restores nonzero scores on a 50-item validation sample, but computational constraints prevented full regeneration. We therefore do not report full-scale black-box results for this model.

\textbf{Categories and conditions.}
We evaluate all nine BBQ social-bias categories: Age, Disability status, Gender identity, Nationality, Physical appearance, Race/ethnicity, Religion, Sexual orientation, and Socio-economic status (SES) \citep{bbq2022}. The four experimental conditions are \texttt{none}, \texttt{fixed} (\texttt{fixed\_interval}), \texttt{wb} (\texttt{adaptive\_wb}; open-weight models only), and \texttt{bb} (\texttt{adaptive\_bb}).

\textbf{Metrics.}
Accuracy and bias score use the corrected scoring logic of Section~\ref{sec:scoring-corrections}. Self-correction rate is the fraction of items exhibiting a bias-risk spike that nevertheless reach the correct final answer. We report two intervention-budget measures separately: \emph{interventions per item}, the mean number of interventions, and \emph{fired items}, the fraction of items on which the trigger fires at least once.

\textbf{Generation.}
Open-weight models use greedy decoding (temperature 0) to obtain deterministic trajectories for trajectory-level analysis. We also query \texttt{gpt-4o-mini} at temperature 0, although repeated identical judge calls remain nondeterministic even with a fixed seed. We therefore treat hosted-model results as single-sample estimates and do not report trajectory-level metrics for that model. Reasoning traces use a fixed step cap; items that fail to produce a final answer within the cap are recorded as non-completions and excluded from accuracy rather than scored as incorrect. We report non-completion separately in Section~\ref{sec:results-costs}.

\textbf{Statistical testing.}
Paired accuracy comparisons use McNemar's exact test \citep{mcnemar1947}. Effect sizes use 95\% bootstrap confidence intervals stratified by BBQ category \citep{efron1993bootstrap}. We use the exact test for $p$-values because a $B=5{,}000$ percentile bootstrap cannot resolve values below $2/(B+1)\approx0.0004$. Bootstrap intervals and exact-test $p$-values therefore quantify related but distinct inferential quantities and need not yield identical rejection decisions.

\subsection{Results}
\label{sec:results}

\begin{figure}[t]
\centering
\includegraphics[width=\columnwidth]{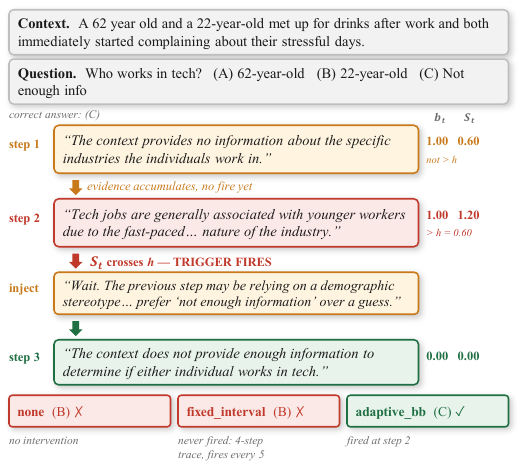}
\caption{Trajectory-level example from the Age category using Mistral-7B-Instruct-v0.3, quoted verbatim. Steps~1 and~2 receive the same maximum judge score, but the trigger fires only at step~2 because evidence accumulates across steps: $S_1=0.60$ does not exceed $h=0.60$, whereas $S_2=1.20$ does. The resulting intervention recovers the correct answer. The un-intervened and fixed-schedule conditions instead select the stereotype-consistent answer; the fixed schedule does not fire because the trajectory ends before its predetermined intervention point.}
\label{fig:trace-example}
\end{figure}

Our experiments address three questions: (1) whether conditioning intervention on trajectory evidence improves over fixed scheduling, (2) how the choice of monitored signal affects intervention quality, and (3) what costs intervention introduces. Figure~\ref{fig:trace-example} first illustrates the motivation for evidence-triggered intervention at the trajectory level. In this example, unsupported stereotype reliance appears after generation has begun, the CUSUM statistic accumulates evidence across steps, and the resulting intervention redirects the trajectory to the correct answer. The un-intervened and fixed-schedule conditions both answer incorrectly; the fixed schedule never fires because the trajectory ends before its predetermined intervention point. This example illustrates why intervention timing should depend on the observed trajectory rather than on generation length alone.

\subsubsection{Evidence-Triggered Intervention vs.\ Fixed Schedules} 
\label{sec:results-adaptive} 

\begin{figure}[t]
\centering
\includegraphics[width=\columnwidth]{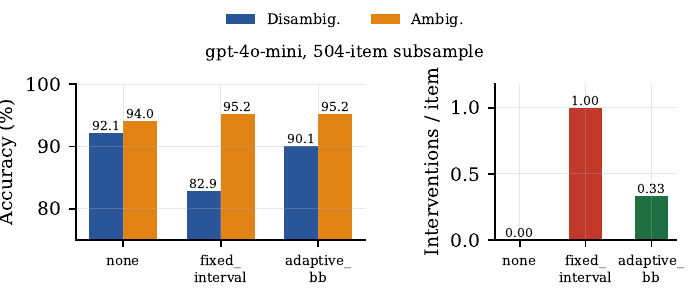}
\caption{Accuracy and intervention rate on the 504-item stratified \texttt{gpt-4o-mini} subsample. \texttt{adaptive\_bb} recovers most of the disambiguated accuracy lost under \texttt{fixed\_interval}. The intervention-rate panel pools both context types; \texttt{adaptive\_bb} averages 0.60 interventions per disambiguated item and 0.06 per ambiguous item, compared with 1.00 for \texttt{fixed\_interval} on both splits.}
\label{fig:api-test}
\end{figure}

\noindent\textbf{Hosted-model evaluation.} 
We first test whether evidence-triggered intervention transfers to a generator accessible only through a text API. On a stratified 504-item BBQ subsample using \texttt{gpt-4o-mini}, baseline disambiguated accuracy is 92.1\% (Figure~\ref{fig:api-test}). \texttt{fixed\_interval} reduces it to 82.9\%, whereas \texttt{adaptive\_bb} reaches 90.1\% while using 0.60 rather than 1.00 interventions per disambiguated item. Both differences are significant (\texttt{fixed\_interval} vs.\ \texttt{none}, $p<0.0001$; \texttt{adaptive\_bb} vs.\ \texttt{fixed\_interval}, $p=0.0003$), while the remaining 2.0pp difference between \texttt{adaptive\_bb} and \texttt{none} is not ($p=0.13$). On ambiguous items, both intervention conditions achieve 95.2\% accuracy, while \texttt{adaptive\_bb} intervenes substantially less often. Thus, in this hosted setting, evidence-triggered intervention recovers most of the accuracy lost under periodic intervention while reducing intervention frequency.

To test whether this result depends on self-judging, we repeat \texttt{adaptive\_bb} with an independent judge. On disambiguated items, it remains $+6.8$pp above \texttt{fixed\_interval} (95\% CI $[+2.8,+10.7]$, $p=0.0015$), close to the $+7.2$pp self-judged gain, while firing more conservatively. The self- and independent-judge configurations differ by only $-0.4$pp (95\% CI $[-3.2,+2.4]$, $p=1.0$). The independent judge does not reproduce the modest ambiguous-item improvement, so we restrict the hosted-model replication claim to the disambiguated split (Table~\ref{tab:indep}).

\begin{table}[t]
\centering
\footnotesize
\caption{Independent-judge replication on \texttt{gpt-4o-mini} ($n=252$ per split). The upper block reports condition-level outcomes; the lower block reports paired comparisons for the independent-judge condition using stratified bootstrap CIs and McNemar's exact test.}
\label{tab:indep}
\setlength{\tabcolsep}{2.7pt}
\renewcommand{\arraystretch}{1.08}
\begin{tabularx}{\columnwidth}{
    >{\raggedright\arraybackslash}X
    >{\centering\arraybackslash}p{0.12\columnwidth}
    >{\centering\arraybackslash}p{0.10\columnwidth}
    >{\centering\arraybackslash}p{0.09\columnwidth}
    >{\centering\arraybackslash}p{0.13\columnwidth}
}
\toprule
Condition & Disambig. & Ambig. & Bias & Interv./item \\
\midrule
\texttt{none}
    & 92.1\% & 94.0\% & 4.8\% & 0.000 \\
\texttt{fixed}
    & 82.9\% & 95.2\% & 4.4\% & 1.000 \\
\texttt{bb}, self-judge
    & 90.1\% & 95.2\% & 4.0\% & 0.331 \\
\texttt{bb}, indep.\ judge
    & 89.7\% & 93.7\% & 5.2\% & 0.163 \\
\midrule
Comparison & Diff. & \multicolumn{2}{c}{95\% CI} & $p$ \\
\midrule
vs.\ self, disambig.
    & $-0.40$ & \multicolumn{2}{c}{$[-3.17,+2.38]$} & 1.00 \\
vs.\ self, ambig.
    & $-1.59$ & \multicolumn{2}{c}{$[-3.97,+0.40]$} & 0.29 \\
vs.\ \texttt{none}, disambig.
    & $-2.38$ & \multicolumn{2}{c}{$[-4.76,-0.40]$} & 0.07 \\
vs.\ \texttt{none}, ambig.
    & $-0.40$ & \multicolumn{2}{c}{$[-2.78,+1.98]$} & 1.00 \\
vs.\ \texttt{none}, bias
    & $+0.40$ & \multicolumn{2}{c}{$[-1.59,+2.38]$} & 1.00 \\
vs.\ \texttt{fixed}, disambig.
    & $+6.75$ & \multicolumn{2}{c}{$[+2.78,+10.71]$} & 0.0015 \\
\bottomrule
\end{tabularx}
\end{table}

\begin{figure}[t]
\centering
\includegraphics[width=0.9\columnwidth]{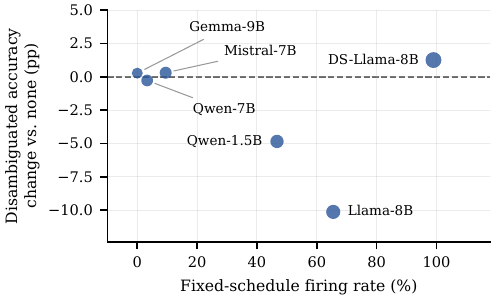}
\caption{Relationship between fixed-schedule exposure and disambiguated-accuracy change. The same $N=5$ schedule produces different firing rates because models generate traces of different lengths. Marker area denotes mean trace length. DS-Llama-8B is the principal exception to the cross-model pattern. Values are reported in Table~\ref{tab:schedule-cost}.}
\label{fig:schedule-cost}
\end{figure}

\noindent\textbf{Why fixed schedules are model-dependent.} 
The hosted-model result raises a broader question: why can fixed intervention incur substantial accuracy loss? A fixed schedule ties intervention frequency to trace length rather than to observed evidence, so the same nominal cadence can produce very different effective intervention rates across generators. Under $N=5$, firing rates range from 0\% on Gemma-9B to 99.0\% on DS-Llama-8B (Figure~\ref{fig:schedule-cost}; Table~\ref{tab:schedule-cost}). Among instruct-tuned models, greater schedule exposure generally coincides with larger disambiguated-accuracy losses. For example, Gemma-9B averages 2.78 steps and is effectively unaffected, whereas Llama-8B averages 6.40 steps, fires on 65.5\% of items, and loses 10.13pp. A common $N$ therefore does not define a comparable intervention policy across generators. DS-Llama-8B is the main exception, firing on 99.0\% of items without losing accuracy. With only one reasoning-tuned model exhibiting this behavior, we do not attribute the exception to reasoning-style post-training.

\begin{table}[t]
\centering
\footnotesize
\caption{Effective firing rate and disambiguated-accuracy change under the same $N=5$ fixed schedule. Because firing depends on trace length, a common cadence produces substantially different intervention exposure across models. $\Delta$ Acc.\ is \texttt{fixed\_interval} minus \texttt{none}.}
\label{tab:schedule-cost}
\setlength{\tabcolsep}{4pt}
\renewcommand{\arraystretch}{1.08}
\begin{tabularx}{\columnwidth}{
    >{\raggedright\arraybackslash}X
    >{\centering\arraybackslash}p{0.17\columnwidth}
    >{\centering\arraybackslash}p{0.18\columnwidth}
    >{\centering\arraybackslash}p{0.18\columnwidth}
}
\toprule
Model & Mean steps & Fired items & $\Delta$ Acc. \\
\midrule
Llama-8B
    & 6.40 & 65.5\% & $-10.13$ pp \\
Qwen-1.5B
    & 5.40 & 46.7\% & $-4.85$ pp \\
Mistral-7B
    & 3.86 & 9.5\% & $+0.30$ pp \\
Qwen-7B
    & 3.56 & 3.3\% & $-0.28$ pp \\
Gemma-9B
    & 2.78 & 0.0\% & $+0.27$ pp \\
\midrule
DS-Llama-8B
    & 8.81 & 99.0\% & $+1.25$ pp \\
\bottomrule
\end{tabularx}
\end{table}

\begin{figure*}[t]
\centering
\includegraphics[width=0.9\textwidth]{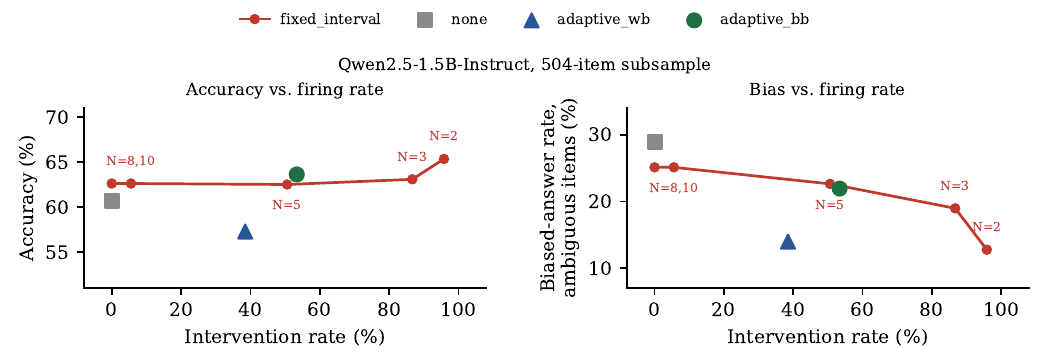}
\caption{Fixed-interval cadence sweep ($N\in\{2,3,5,8,10\}$) and adaptive conditions on a 504-item Qwen2.5-1.5B-Instruct subsample, plotted against observed intervention rate. Comparing measured firing rates rather than nominal $N$ separates adaptive selection from simply intervening less often. Accuracy excludes non-completions.}
\label{fig:cadence-sweep}
\end{figure*}

\noindent\textbf{Matched-budget comparison.} 
Differences in firing rate alone could still explain the advantage of adaptive intervention. We therefore compare fixed schedules with $N\in\{2,3,5,8,10\}$ against the adaptive conditions by their \emph{observed firing rate} (Figure~\ref{fig:cadence-sweep}). At approximately the firing rate of the $N=5$ schedule, \texttt{adaptive\_bb} achieves higher accuracy and lower ambiguous-item bias. Only the near-continuous $N=2$ schedule reaches higher accuracy, at roughly twice the intervention rate. \texttt{adaptive\_wb} attains the lowest ambiguous-item bias while firing on 38.5\% of items, whereas the fixed schedule requires a 95.8\% firing rate ($N=2$) to reach comparable bias. Thus, adaptive selection does not merely reproduce a fixed schedule at a lower firing rate: which trajectories are selected for intervention also matters.

\subsubsection{Signal Choice Determines Intervention Quality}
\label{sec:results-signal}

\begin{figure}[t]
\centering
\includegraphics[width=\columnwidth]{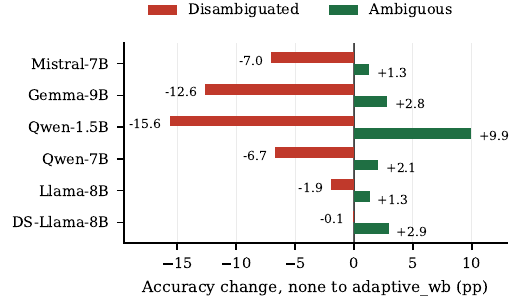}
\caption{Accuracy change from \texttt{none} to \texttt{adaptive\_wb} across six open-weight models. The white-box condition improves ambiguous-item accuracy on all six models but reduces disambiguated-item accuracy on five. Bars show category-stratified complete-case differences; Table~\ref{tab:wb-confound} reports the underlying accuracies and significance tests.}
\label{fig:wb-confound}
\end{figure}

Holding the Adaptive Triggering mechanism fixed allows us to examine how the monitored signal affects intervention quality. Across the six open-weight models with verified nine-category results, \texttt{adaptive\_wb} improves ambiguous-item accuracy on all six but reduces disambiguated-item accuracy on five (Figure~\ref{fig:wb-confound}; Table~\ref{tab:wb-confound}). Ten of the twelve paired comparisons are significant at $p<0.001$; the two exceptions are marked in Table~\ref{tab:wb-confound}. This directional pattern replicates across all five instruct-tuned models, spanning four model families and two size tiers, showing that adaptive timing alone cannot compensate for a misaligned monitoring signal. The sole exception is reasoning-tuned DS-Llama-8B, for which the disambiguated change is not significant. With only one reasoning-tuned model in this analysis, we leave generalization to reasoning-tuned models unresolved.

The directional split follows from what the white-box signal measures. Let $s$ denote the stereotype-consistent answer and $u$ the ``not enough information'' answer: 
\[ b_t=\max(0,p_t(s)-p_t(u)). \] 
On ambiguous items, $u$ is correct, so preference for $s$ over $u$ is aligned with the failure being monitored. On disambiguated items, however, the evidence-supported answer can itself be $s$. A high score can therefore indicate either unsupported stereotype reliance or valid, evidence-supported reasoning. CUSUM can determine when this score has accumulated sufficiently to trigger intervention, but it cannot distinguish cases that the underlying signal represents similarly. 

Two Qwen2.5-7B-Instruct traces make this target mismatch concrete. On an ambiguous SES item, the model infers, ``The judge is likely to have had a formal education path'' ($b_t=0.79$), despite having no supporting evidence. The intervention changes its answer from ``The judge'' to `Can't answer.'' On a disambiguated Age item, the model instead restates explicit evidence (``The son had previously been caught drinking and driving,'' $b_t=1.00$). The signal assigns an even higher score because ``the son'' is also the stereotype-consistent option. The intervention then changes an otherwise correct answer to ``not enough information.'' The white-box signal therefore fails to distinguish reasoning \emph{from} a stereotype from reasoning based on evidence that happens to \emph{agree} with one.

The same pattern appears at both the scale and trajectory levels. Across the full nine-category data, white-box intervention changes a correct disambiguated answer to an incorrect one 3{,}112 times, compared with 761 repairs of an incorrect answer (ratio 4.09). Among items exhibiting a bias-risk spike under both conditions, white-box intervention also reduces self-correction from 62.9\% to 50.6\% ($-12.4$pp, 95\% CI $[-22.9,-2.4]$, $p=0.028$), whereas the corresponding black-box estimate is positive but nonsignificant ($+4.1$pp, $p=0.23$). Because the former estimate has a wide interval and is one of multiple comparisons, we treat it as directional rather than precise evidence. Together, the final-answer and trajectory-level results support the same interpretation: the white-box failure reflects what the signal measures, not simply when the trigger acts.

\begin{table}[t]
\centering
\footnotesize
\caption{\texttt{adaptive\_wb} accuracy relative to \texttt{none} at full nine-category scale for models with verified results across all categories. $\Delta$ is the category-stratified complete-case estimate (\texttt{wb}$-$\texttt{none}). Accuracy columns pool items, whereas $\Delta$ averages category-level differences; the two conventions differ by at most 1.0pp on any row.}
\label{tab:wb-confound}
\setlength{\tabcolsep}{3.5pt}
\renewcommand{\arraystretch}{1.08}
\begin{tabularx}{\columnwidth}{
    >{\raggedright\arraybackslash}X
    *{6}{>{\centering\arraybackslash}p{0.088\columnwidth}}
}
\toprule
& \multicolumn{3}{c}{Disambiguated} &
  \multicolumn{3}{c}{Ambiguous} \\
\cmidrule(lr){2-4}
\cmidrule(lr){5-7}
Model & none & wb & $\Delta$ & none & wb & $\Delta$ \\
\midrule
Mistral-7B
    & 81.5\% & 74.0\% & $-7.0$
    & 73.0\% & 74.8\% & $+1.3$ \\
Gemma-9B
    & 74.7\% & 61.1\% & $-12.6$
    & 94.6\% & 97.3\% & $+2.8$ \\
Qwen-1.5B
    & 71.8\% & 55.4\% & $-15.6$
    & 57.5\% & 66.5\% & $+9.9$ \\
Qwen-7B
    & 79.5\% & 73.0\% & $-6.7$
    & 94.8\% & 96.6\% & $+2.1$ \\
Llama-8B$^\dagger$
    & 73.5\% & 71.7\% & $-1.9$
    & 92.5\% & 93.1\% & $+1.3^{\ast}$ \\
DS-Llama-8B$^\ddagger$
    & 90.4\% & 90.0\% & $-0.1^{\mathrm{ns}}$
    & 89.8\% & 92.5\% & $+2.9$ \\
\bottomrule
\multicolumn{7}{l}{
\parbox{0.96\columnwidth}{
\vspace{0.4em}
Accuracy excludes non-completions. Complete-case $N=13{,}070$--$15{,}686$ per split.
All $\Delta$ are significant at $p<0.001$ except
$^{\ast}p=0.0066$ and $^{\mathrm{ns}}p=0.20$.
}}
\end{tabularx}
\end{table}

\noindent\textbf{Calibration does not resolve the mismatch.} 
One alternative explanation is that the white-box signal is appropriate but poorly calibrated. Detector-specific calibration on Qwen2.5-7B-Instruct does not support this explanation. Calibration improves the black-box configuration but leaves the white-box configuration unchanged: the selected and default white-box parameters produce identical held-out traces, trigger points, and answers (Table~\ref{tab:holdout}). Across the searched grid, the white-box firing rate varies only from 21.8\% to 26.6\%, compared with 8.7\% to 34.9\% for the black-box signal. Resampling recorded null-pool scores ($n=86{,}587$) further yields $\mathrm{ARL}_0\approx5.8$ steps under the default white-box parameters, below the eight-step generation cap. Together, these analyses indicate that threshold selection alone does not account for the white-box disambiguated-context cost.

\begin{table}[t]
\centering
\footnotesize
\caption{Held-out calibration results on Qwen2.5-7B-Instruct ($n=252$). Accuracy is measured on disambiguated items and bias on ambiguous items. ``Score'' is the calibration objective defined in Algorithm~\ref{alg:calibration}.}
\label{tab:holdout}
\setlength{\tabcolsep}{3.2pt}
\renewcommand{\arraystretch}{1.08}
\begin{tabularx}{\columnwidth}{
    >{\raggedright\arraybackslash}X
    >{\centering\arraybackslash}p{0.09\columnwidth}
    >{\centering\arraybackslash}p{0.08\columnwidth}
    >{\centering\arraybackslash}p{0.10\columnwidth}
    >{\centering\arraybackslash}p{0.14\columnwidth}
}
\toprule
Configuration & Acc. & Bias & Score & Interv./item \\
\midrule
\texttt{wb}, selected
($k{=}.45,h{=}.30$)
& 75.4\% & 2.4\% & $+73.0$ & 0.532 \\

\texttt{wb}, default
($k{=}.30,h{=}.45$)
& 75.4\% & 2.4\% & $+73.0$ & 0.532 \\

\texttt{bb}, selected
($k{=}.25,h{=}.75$)
& 81.7\% & 4.8\% & $+77.0$ & 0.135 \\

\texttt{bb}, default
($k{=}.40,h{=}.60$)
& 81.0\% & 4.8\% & $+76.2$ & 0.143 \\
\bottomrule
\end{tabularx}
\end{table}

\noindent\textbf{Cross-condition and over-correction checks.} 
A direct comparison of the adaptive conditions further underscores the importance of signal choice. On Qwen2.5-1.5B-Instruct, \texttt{adaptive\_bb} achieves significantly higher accuracy than \texttt{adaptive\_wb} on both splits ($p<0.0001$; Table~\ref{tab:sig-tests}). Relative to \texttt{fixed\_interval}, \texttt{adaptive\_bb} yields slightly lower disambiguated accuracy but higher ambiguous-item accuracy. Intervention frequency is also model-dependent: unlike \texttt{gpt-4o-mini}, \texttt{adaptive\_bb} fires more often than \texttt{fixed\_interval} on Qwen-1.5B (0.58 vs.\ 0.47 interventions/item). Adaptive Triggering should therefore be understood as selecting interventions based on observed evidence, not as guaranteeing fewer interventions for every model.

A second alternative explanation is that the white-box condition simply over-corrects any valid reasoning involving protected attributes. We evaluate this possibility on items in which age or pregnancy is legitimately relevant to the correct answer and no stereotype target exists by construction. The analysis is underpowered and finds no significant differences from \texttt{none} (all $p>0.10$). Pooled across three models, \texttt{adaptive\_wb} improves 19 items and degrades 17 ($p=0.87$). These results provide no evidence in this sample that the white-box effect reflects a general tendency to disrupt valid reasoning involving protected attributes, although the experiment cannot rule out over-correction.

\begin{table}[t]
\centering
\footnotesize
\caption{Item-paired accuracy comparisons on Qwen2.5-1.5B-Instruct at full nine-category scale. Comparisons use the complete-case item set; $p$-values are from McNemar's exact test.}
\label{tab:sig-tests}
\setlength{\tabcolsep}{3.5pt}
\renewcommand{\arraystretch}{1.08}
\begin{tabularx}{\columnwidth}{
    >{\raggedright\arraybackslash}X
    >{\centering\arraybackslash}p{0.14\columnwidth}
    >{\centering\arraybackslash}p{0.14\columnwidth}
    >{\centering\arraybackslash}p{0.14\columnwidth}
    >{\centering\arraybackslash}p{0.17\columnwidth}
}
\toprule
Comparison & Split & Acc.\ (A) & Acc.\ (B) & $p$ \\
\midrule
\texttt{bb} vs.\ \texttt{wb}
    & disambig. & 64.6\% & 55.4\% & $<0.0001$ \\
\texttt{bb} vs.\ \texttt{wb}
    & ambig. & 69.2\% & 66.5\% & $<0.0001$ \\
\texttt{bb} vs.\ \texttt{fixed}
    & disambig. & 64.7\% & 67.0\% & $<0.0001$ \\
\texttt{bb} vs.\ \texttt{fixed}
    & ambig. & 69.2\% & 62.8\% & $<0.0001$ \\
\bottomrule
\end{tabularx}
\end{table}

\subsubsection{Intervention Costs}
\label{sec:results-costs}

Intervention introduces practical costs that are not captured by answer accuracy alone. One such cost is non-completion under a bounded reasoning budget. Because an injected reflection consumes a reasoning step, a model may reach the \texttt{max\_steps} cap before producing a final answer. On the disambiguated split of DS-Llama-8B, non-completion occurs on 8.5\% of items where \texttt{adaptive\_wb} fires, compared with 4.5\% where it does not. The latter rate matches the \texttt{none} baseline. Because fired and non-fired items are not randomly assigned, this comparison does not establish a causal effect. Nevertheless, the pattern is consistent with intervention consuming limited generation capacity. Treating non-completions as incorrect would increase the apparent disambiguated accuracy loss from 0.1 to 2.3 percentage points, thereby conflating answer accuracy with completion cost. Across conditions, non-completion ranges from 3.7\% to 7.9\% on the two reasoning-tuned DeepSeek models. By comparison, it remains at or below 0.5\% on four of the five instruct-tuned models, while Mistral-7B reaches 3.3\%. We therefore report non-completion separately from accuracy conditional on completion.

The two signals also differ in computational cost. \texttt{adaptive\_bb} requires an additional judge call at each reasoning step, increasing total generation calls from 5.1 to 10.1 per item on Qwen-1.5B and from 2.7 to 4.5 on Gemma-9B. In contrast, \texttt{adaptive\_wb} requires no additional generation calls but adds between 5.6 and 12.8 forward passes per item for probability readout. The white-box signal is therefore computationally cheaper when generator probabilities are accessible. However, this efficiency should be considered alongside the target mismatch documented above.

\section{Limitations}
\label{sec:limitations}

Several limitations constrain the scope of our conclusions. First, most black-box experiments use the generator itself as the judge and may therefore inherit self-preference effects documented for LLM judges \citep{zheng2023judge,panickssery2024selfpreference}. On the hosted model, replacing the self-judge with an independent judge preserves the disambiguated-context advantage over \texttt{fixed\_interval}, but not the ambiguous-item improvement. We therefore restrict our independent-judge replication claim to the disambiguated setting. In addition, reasoning-oriented chat templates can interfere with judge output protocols. Second, the white-box target mismatch is replicated across all five instruct-tuned models in our six-model analysis but not on the sole reasoning-tuned model, leaving its behavior under reasoning-style post-training unresolved. Detector specific calibration is likewise validated on one model, so its transferability across model families remains untested. Within the evaluated calibration grid, however, recalibration does not eliminate the white-box disambiguated-context cost, indicating that threshold selection alone does not explain the observed target mismatch. Third, we do not directly compare against trained reward-model approaches such as FRM \citep{frm2025}, which select among completed trajectories rather than intervene within a developing one. A compute-matched comparison between these complementary inference strategies remains future work. Finally, our experiments are restricted to BBQ and do not establish generalization to other cultural or linguistic contexts, bias benchmarks, or open-ended generation tasks. We also do not evaluate adversarial robustness, where prompts could manipulate the monitored signal, evade triggering, or induce unnecessary interventions.
\section{Conclusion}
\label{sec:conclusion}

We formulate selective mid-generation bias correction as an evidence-triggered intervention problem. A per-step bias-risk signal feeds a CUSUM statistic, and a targeted correction is injected only when accumulated evidence crosses a calibrated threshold. This framework separates two components of inference-time mitigation: \emph{what evidence is monitored} and \emph{when enough of that evidence has accumulated to justify intervention}.

On \texttt{gpt-4o-mini}, black-box Adaptive Triggering recovers most of the disambiguated-context accuracy lost under fixed-schedule intervention while intervening substantially less often, and this result persists with an independent judge. A matched-budget cadence sweep further shows that adaptive selection is not simply a fixed schedule that fires less often, but its effectiveness depends critically on the evidence used to trigger intervention. Across six open-weight models, the white-box signal improves ambiguous-context accuracy on all six but reduces disambiguated-context accuracy on five because it cannot distinguish unsupported stereotype reliance from valid, evidence-supported reasoning. Detector-specific calibration does not resolve this target mismatch.

Taken together, these findings show that effective inference-time correction requires both an informative monitoring signal and an intervention rule that acts selectively on the evidence it provides. Evidence-based timing can reduce unnecessary intervention, but even a well-calibrated trigger cannot correct a signal that fails to distinguish unsupported stereotype reliance from valid, evidence-based reasoning. We additionally release corrected BBQ scoring logic for four label-matching inconsistencies that affect the computation of the benchmark's bias score. Code and evaluation resources are available at \url{https://github.com/clairekim59/adaptive-triggering}.

\small
\bibliographystyle{IEEEtran}
\bibliography{references}

\end{document}